\pdfoutput=1

\documentclass[11pt]{article}

\usepackage[final]{acl}

\usepackage{times}
\usepackage{latexsym}

\usepackage[T1]{fontenc}

\usepackage[utf8]{inputenc}

\usepackage{microtype}

\usepackage{inconsolata}

\usepackage[inline]{enumitem}

\usepackage{graphicx}
\usepackage{algorithm}
\usepackage{algorithmic}
\usepackage{xcolor}
\usepackage[inline]{enumitem}
\usepackage{multirow}
\usepackage[T1]{fontenc}
\usepackage{pgfplots}
\usepackage{tikz}
\usepackage{amsmath}
\usepackage{makecell}
\definecolor{bblue}{HTML}{4F81BD}
\definecolor{rred}{HTML}{FFB303}
\definecolor{ggreen}{HTML}{9BBB59}
\definecolor{igreen}{HTML}{579c35}
\definecolor{ppurple}{HTML}{9F4C7C}

\pgfplotsset{
   compat=1.14,
   legend entry/.initial=,
   every axis plot post/.code={%
       \pgfkeysgetvalue{/pgfplots/legend entry}\tempValue
       \ifx\tempValue\empty
           \pgfkeysalso{/pgfplots/forget plot}%
       \else
           \expandafter\addlegendentry\expandafter{\tempValue}%
       \fi
   },
}

\usepackage{newfloat}
\usepackage{listings}
\DeclareCaptionStyle{ruled}{labelfont=normalfont,labelsep=colon,strut=off} 
\floatstyle{ruled}
\newfloat{listing}{tb}{lst}{}
\floatname{listing}{Listing}
\usepackage{longtable}
\usepackage{array}
\usepackage{booktabs}
\usepackage{tabularx}
\usepackage{textcomp}

\title{LLMs are not Zero-Shot Reasoners for Biomedical Information Extraction}

\author{
Aishik Nagar\textsuperscript{1} \quad 
Viktor Schlegel\textsuperscript{2,3} \quad 
Thanh-Tung Nguyen \quad 
Hao Li\textsuperscript{3} \quad \\
\textbf{Yuping Wu}\textsuperscript{3} \quad 
\textbf{Kuluhan Binici}\textsuperscript{4} \quad 
\textbf{Stefan Winkler}\textsuperscript{4} \\
\\
\textsuperscript{1}ASUS Intelligent Cloud Services (AICS), Singapore \\
\textsuperscript{2}Imperial College London, Imperial Global Singapore \\
\textsuperscript{3}University of Manchester, United Kingdom \\
\textsuperscript{4}National University of Singapore, Singapore \\
Correspondence to: \texttt{aishik\_nagar@asus.com}
}

\begin{document}

\maketitle

\begin{abstract}
Large Language Models (LLMs) are increasingly adopted for applications in healthcare, reaching the performance of domain experts on tasks such as question answering and document summarisation. Despite their success on these tasks, it is unclear how well LLMs perform on tasks that are traditionally pursued in the biomedical domain, such as structured information extraction. To bridge this gap, in this paper, we systematically benchmark LLM performance in Medical Classification and Named Entity Recognition (NER) tasks. We aim to disentangle the contribution of different factors to the performance, particularly the impact of LLMs' task knowledge and reasoning capabilities, their (parametric) domain knowledge, and addition of external knowledge. To this end, we evaluate various open LLMs---including BioMistral and Llama-2 models---on a diverse set of biomedical datasets, using standard prompting, Chain-of-Thought (CoT) and Self-Consistency based reasoning as well as Retrieval-Augmented Generation (RAG) with PubMed and Wikipedia corpora. Counter-intuitively, our results reveal that standard prompting consistently outperforms more complex techniques across both tasks, laying bare the limitations in the current application of CoT, self-consistency and RAG in the biomedical domain. Our findings suggest that advanced prompting methods developed for knowledge- or reasoning-intensive tasks, such as CoT or RAG, are not easily portable to biomedical tasks where precise structured outputs are required. This highlights the need for more effective integration of external knowledge and reasoning mechanisms in LLMs to enhance their performance in real-world biomedical applications.
\end{abstract}

\section{Introduction}

The success of Large Language Models (LLMs) is reshaping AI healthcare applications, particularly in Question Answering~\cite{DBLP:journals/widm/BudlerGS23,Subramanian2024M-QALM:Answering}, summarization~\cite{van2024adapted,DBLP:conf/clef/SchlegelLW0NKBZ23,nagar2024umedsum}, and extracting insights from unstructured patient-generated health data~\cite{DBLP:conf/bionlp/LiWSBNKZB0N23}. While advancements in fine-tuning and in-context learning (ICL) have improved LLM performance, these rely on readily available structured training data~\cite{DBLP:conf/sepln/AbburiSPVBB23,DBLP:conf/icassp/ZhangW0Z24, DBLP:conf/emnlp/GutierrezMWCLS022}. However, in biomedical contexts, such resources are often unavailable due to domain shifts \cite{hadi2023survey} or ad-hoc requirements —for example when researchers need to process a set of medical records to find patients satisfying inclusion criteria for a clinical trial ~\cite{Jullien2023NLI4CT:Reports,hadi2023survey} (e.g., whether they’re a smoker). This limits the effectiveness of parametric knowledge improvements in LLMs, necessitating strong zero-shot capabilities for structured prediction tasks such as biomedical classification and Named Entity Recognition. Despite this, the literature currently lacks a systematic investigation of other crucial aspects of knowledge utilization in zero-shot performance of LLMs in such tasks.

In order to address this research gap, we first postulate that LLM performance in true zero-shot settings---where only task labels and their meaningful names are provided~\cite{DBLP:journals/pami/LampertNH14}---hinges on three categories of knowledge:
\begin{enumerate*}[label=\emph{(\alph*)}]
\item \textit{Parametric Knowledge}: Information embedded in model weights;
\item \textit{Task Knowledge}: Understanding of task-specific labels and context;
\item \textit{External Knowledge}: Additional retrieved context to supplement decision-making.
\end{enumerate*}

Existing research evaluating these factors in LLMs for the medical domain focus on knowledge-intensive tasks like Multiple-Choice QA~\cite{DBLP:journals/corr/abs-2311-16452,Subramanian2024M-QALM:Answering}, but their capabilities in structured prediction tasks, such as medical classification and information extraction, remain underexplored. Additionally, techniques like zero-shot Chain-of-Thought (CoT) reasoning~\cite{wei2022chain, wang2024chain}, self-consistency~\cite{wang2022self}, and Retrieval-Augmented Generation (RAG)~\cite{li2024rt} require systematic evaluation in these contexts.

Additionally, evaluations often focus on proprietary models like ChatGPT~\cite{biswas2023role} or GPT-4~\cite{DBLP:journals/corr/abs-2303-08774}, which pose challenges due to computational cost, privacy concerns, and inaccessibility for techniques like constrained decoding. Despite the growing concerns regarding reliability of LLMs in medical applications, techniques like constrained decoding which have shown promise in mitigating LLM hallucinations have not been systematically applied to medical information extraction or classification. 

Thus, four key issues currently hinder progress:
\begin{enumerate*}[label=\emph{(\roman*)}]
\item Reliance on training sets and parametric knowledge for structured prediction, which may be unrealistic;
\item Lack of true zero-shot evaluations for structured tasks beyond surrogate QA;
\item Dependence on large-scale, proprietary LLMs, limiting practical deployment.
\item Lack of a systematic analysis of impact of latest techniques such as Chain-of-thought reasoning, RAG and constrained generation in medical structured prediction tasks.
\end{enumerate*}

This paper systematically benchmarks LLMs in medical classification and Named Entity Recognition (NER), assessing task and external knowledge while controlling parametric knowledge. We evaluate CoT reasoning, RAG, and constrained generation, offering insights into their applicability.

\textbf{First}, we present the first comprehensive benchmark of task and external knowledge adaptation for LLMs in medical structured prediction tasks. \textbf{Second}, we analyze the impact of knowledge enhancement techniques, including CoT, self-consistency, RAG, and constrained generation. \textbf{Third}, we demonstrate that parametric knowledge capacity, i.e., model size, is the primary driver of zero-shot performance, highlighting the limitations and potential of current LLM architectures.

\section{Related Work}
We survey the existing benchmarking literature for the medical domain in the \textbf{appendix section \ref{sec:appendix-rel-work}}, outlining the lack of studies focusing on structured prediction tasks. 
Furthermore, we cover recent prompting techniques that were proposed to elicit reasoning in LLMs, and augment their domain knowledge, either by better tapping into their parametric knowledge or by explicitly providing them with relevant external context. 
Notably, we omit approaches that rely on existence of training sets, such as few-shot prompting \cite{DBLP:conf/adma/WangPSCSML23} or model fine-tuning, as one of the key challenges in the medical domain is the lack of annotated task data, due to privacy concerns over sharing medical records. Instead, as outlines in the introduction, we focus on ``true'' zero-shot capabilities of LLMs. 

\textbf{Reasoning- and Knowledge-enhancing approaches:} Current work attempts to improve the performance of LLMs from different knowledge utilization perspectives. One of the obvious methods is full parameter domain-specific pre-training~\cite{xie2024me}. For example,~\citet{chen2023meditron} propose the largest medical foundation model, trained on both biomedical and clinical data, up to 70B. \citet{bolton2024biomedlm}, on the other hand, believe larger LLMs are computationally expensive to run, proposing a 2.7B LLM specific for biomedical NLP tasks. When fine-tuned, the relatively small model compete with larger LLMs. 
In our study, we compare domain-generalist models with those adapted to the medical domain.
Since full parameter tuning is costly, many works focus on domain knowledge adaptation by pre-training \cite{shi2024medadapterefficienttesttimeadaptation, song2024llm} or instruction tuning \cite{willard2023efficient} with adapters. Training-free approaches encompass chain-of-thought (CoT) \cite{wei2022chain,jeong2024olaph}, self-consistency~\cite{wang2022self}, and,  
concerned with lack of grounding resulting in hallucination, recent work introduce RAG methods~\cite{li2024rt, wang2024rat, yu2023retrieval, munnangi2024fly,wang2024jmlr, soong2023improving}. However, most of these efforts have focused on performance in a particular knowledge paradigm and have lacked a systematic assessment of their performance on structured prediction, which we address in our study.

\section{Methodology}



Our methodology is designed to answer the following two research questions:

\emph{1. How well do Large Language Models (LLMs) perform on structured prediction tasks when provided with unstructured inputs?}

\emph{2. To what extent can approaches that enhance task knowledge and external knowledge improve their performance?}
To answer the first research question, we benchmark LLMs on biomedical text classification and NER datasets, focusing on the ``true'' zero-shot setting to evaluate models' \textit{parametric knowledge}. This reflects real-world scenarios where annotated data is often unavailable due to ad-hoc task requirements, resource limitations and privacy constraints \citep{giachelle2021medtag}. 
This leads to what~\citet{fries2022dataset} describe as ``dataset debt'', highlighting issues like inconsistent documentation, lack of domain-specific information except generic entities and difficulties adapting datasets to niche domains.
Clinicians face significant time constraints, which limit even few-shot annotations \citep{xia2012clinical, wac2024evaluation, farri2013effects}. 
These factors make \textbf{fine-tuning and few-shot approaches impractical} for structured prediction tasks in the biomedical domain, \textbf{positioning zero-shot methods as a scalable solution for real-world biomedical tasks}.

To answer the second question, we compare their zero-shot performance to various methods that aim to enhance \textit{task knowledge} and \textit{external knowledge}, while keeping the \textit{parametric knowledge} static.

\paragraph{Techniques} Table~\ref{tab:techniques_summary} lists our methods. We use \textsc{Vanilla} prompting as the baseline, and enhance it with advanced approaches: chain-of-thought (\textsc{CoT}) \cite{wei2022chain} and self-consistency (\textsc{SC}) \cite{wang2022self}, along with retrieval-augmented generation (\textsc{RAG}) \cite{lewis2020retrieval} that leverages FAISS with PubMed abstracts and Wikipedia articles, embedding documents via \texttt{all-MiniLM-L6-v2} \cite{DBLP:conf/emnlp/ReimersG19}. We also apply constrained decoding \cite{willard2023efficient} to enforce structured outputs. For NER, we adopt a two-stage approach: Stage~1 assigns generic entity labels (e.g., “Bodypart”), and Stage~2 refines them to fine-grained labels. Self-consistency is employed in both tasks to aggregate multiple reasoning paths via majority voting.

\begin{table*}[!t]
\centering
\footnotesize
\begin{tabular}{p{0.47\columnwidth}p{0.43\columnwidth}p{1\columnwidth}}
\hline
\textbf{Technique} & \textbf{Details} & \textbf{Comments} \\ \hline
\textsc{Vanilla} & Standard prompting. & Baseline for all tasks. \\ \hline
\textsc{Chain-of-Thought (CoT)} \cite{wei2022chain} & Chain-of-thought reasoning. & Effective for QA and logical reasoning. For NER, adapted into a two-stage approach where generic entity names are first induced (e.g., \textit{Bodypart}), followed by fine-grained labeling. \\ \hline
\textsc{Self-Consistency (SC)} \cite{wang2022self} & Majority voting across sampled reasoning paths. & Applied in both stages of the two-stage NER approach. \\ \hline
\textsc{Retrieval-Augmented-Generation (RAG)} \cite{lewis2020retrieval} & Retrieval-augmented generation using FAISS \cite{douze2024faiss}. & Used PubMed \cite{DBLP:journals/jis/SanyalBD21} and Wikipedia as corpora. PubMed improved performance; Wikipedia degraded performance for medical QA \cite{xiong2024benchmarking}. \\ \hline
\textsc{Constrained Decoding} \cite{willard2023efficient} & Restricted outputs to ensure structured extraction. & Avoided hallucinations. Ensured span and label consistency in NER tasks. \\ \hline
\end{tabular}
\caption{Techniques Summary with Comments and Details. Complete details can be found in Appendix \ref{sec:techniques-full}.}
\label{tab:techniques_summary}
\end{table*}

\textbf{Complete details of our datasets, techniques and methods are described in Appendix \ref{sec:techniques-full}}.

\section{Evaluation Results}
The complete table of results is provided in Table~\ref{table:combined_results_with_span}. We give an overview of our findings followed by a deeper analysis of the evaluated techniques.

\subsection{Overview of results}

\begin{table}[H]
\centering
\begin{tabular}{llrrr}
\toprule
\multirow{2}{*}{} & \multirow{2}{*}{Technique} & \multicolumn{1}{c}{CLS} & \multicolumn{2}{c}{NER} \\
\cmidrule(lr){3-3} \cmidrule(lr){4-5}
&  & F1 & F1-S & F1-L \\
\midrule
\multirow{7}{*}{\rotatebox[origin=c]{90}{\footnotesize{BioMistral-7B}}} & \textsc{Vanilla} & 36.5 & 3.3 & 2.2 \\
& \textsc{CoT} & 31.3 & 1.5 & 1.3 \\
& \textsc{SC-CoT} & 20.5 & 0.8 & 0.4 \\
& \textsc{CoT-RAG-P} & 14.7 & 1.6 & 1.2 \\
& \textsc{CoT-RAG-W} & 15.5 & 1.3 & 1.0 \\
& \textsc{SC-CoT-RAG-P}& 19.2 & 0.5 & 0.4 \\
& \textsc{SC-CoT-RAG-W}& 21.6 & 0.4 & 0.3 \\
\midrule
\multirow{7}{*}{\rotatebox[origin=c]{90}{\footnotesize{Llama-2-70B}}} & \textsc{Vanilla} & 40.3 & 8.6 & 5.8 \\
& \textsc{CoT} & 35.9 & 10.3 & 7.3 \\
& \textsc{SC-CoT} & 28.0 & 9.1 & 5.4 \\
& \textsc{CoT-RAG-P} & 16.5 & 9.9 & 7.1 \\
& \textsc{CoT-RAG-W} & 15.7 & 10.6 & 7.2 \\
& \textsc{SC-CoT-RAG-P}& 27.2 & 9.0 & 5.4 \\
& \textsc{SC-CoT-RAG-W}& 26.6 & 9.1 & 5.3 \\
\midrule
\multirow{7}{*}{\rotatebox[origin=c]{90}{\footnotesize{Llama-2-7B}}} & \textsc{Vanilla} & 34.9 & 6.5 & 5.2 \\
& \textsc{CoT} & 30.6 & 4.9 & 2.5 \\
& \textsc{SC-CoT} & 24.6 & 5.1 & 3.0 \\
& \textsc{CoT-RAG-P} & 14.3 & 4.6 & 2.3 \\
& \textsc{CoT-RAG-W} & 14.5 & 4.2 & 1.7 \\
& \textsc{SC-CoT-RAG-P}& 25.5 & 5.7 & 2.9 \\
& \textsc{SC-CoT-RAG-W}& 11.1 & 5.6 & 3.2 \\
\bottomrule
\end{tabular}
\caption{Performance of each model and technique combination across Classification and NER datasets. For classification, we report Micro-F1 and for NER we report both Span-Identification Micro-F1 performance as well as full Micro-F1 performance, including recognizing correct types.}
\label{table:combined_results_with_span}
\end{table}

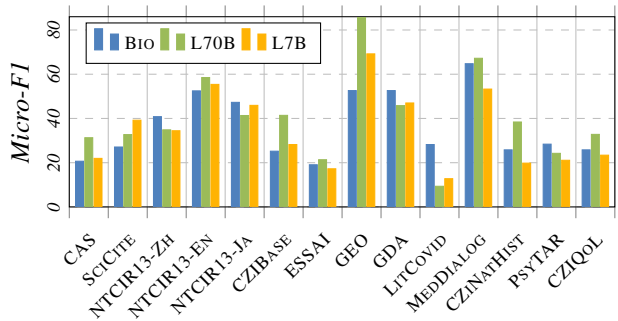
\begin{figure}[t]
\centering
    \begin{tikzpicture}
    \begin{axis}[
        ybar=0.5pt,
        bar width=3pt,
        x tick label style={rotate=45, font=\scriptsize\scshape, anchor=north east},
        x tick label as interval,
        y tick label style={rotate=90, font=\scriptsize\itshape},
        ytick={0, 0.2, 0.4, 0.6, 0.8},
        yticklabels={0, 20, 40, 60, 80},
        height=11.7em,
ylabel=\emph{Micro-F1},
        width  = 1.05\columnwidth,
        ymajorgrids=true,
        y grid style=dashed,
        major y tick style = transparent,
        xtick={0, 1, 2, 3, 4, 5, 6, 7, 8, 9, 10, 11, 12, 13, 14},
        xmajorgrids=true,
        xminorgrids=true,
        legend pos=north west,
        legend cell align={left},
        legend columns=3,
        xmin=0, xmax=14,
        ymin=0.0, ymax=0.86,
        xticklabels={CAS, SciCite, NTCIR13-Zh, NTCIR13-En, NTCIR13-Ja, CZIBase, ESSAI, GEO, GDA, LitCovid, MedDialog, CZiNatHist, PsyTAR, CZIQoL}
    ]
    \addplot[legend entry=\textsc{\scriptsize Bio}, color=bblue, fill=bblue]  coordinates {(0.5, 0.2066) (1.5, 0.2709) (2.5, 0.4081) (3.5, 0.5247) (4.5, 0.4728) (5.5, 0.2520) (6.5, 0.1906) (7.5, 0.5260) (8.5, 0.5260) (9.5, 0.2821) (10.5, 0.6475) (11.5, 0.2580) (12.5, 0.2838) (13.5, 0.2580)};
    \addplot[legend entry=\textsc{\scriptsize L70B}, color=ggreen, fill=ggreen]  coordinates {(0.5, 0.3129) (1.5, 0.3270) (2.5, 0.3485) (3.5, 0.5849) (4.5, 0.4133) (5.5, 0.4140) (6.5, 0.2134) (7.5, 0.8560) (8.5, 0.4580) (9.5, 0.0934) (10.5, 0.6721) (11.5, 0.3840) (12.5, 0.2423) (13.5, 0.3280)};
    \addplot[legend entry=\textsc{\scriptsize L7B}, color=rred, fill=rred]  coordinates {(0.5, 0.2195) (1.5, 0.3921) (2.5, 0.3444) (3.5, 0.5540) (4.5, 0.4592) (5.5, 0.2820) (6.5, 0.1724) (7.5, 0.6920) (8.5, 0.4700) (9.5, 0.1276) (10.5, 0.5328) (11.5, 0.1980) (12.5, 0.2109) (13.5, 0.2340)};
    \end{axis}
    \end{tikzpicture}
    \vspace{-.33\baselineskip}
    \caption{Best-performing \emph{Standard Prompting} method for \textcolor{bblue}{\bfseries Bio}Mistral 7B, \textcolor{ggreen}{\bfseries L}lama-\textcolor{ggreen}{\bfseries 70B} and \textcolor{rred}{\bfseries L}lama-\textcolor{rred}{\bfseries 7B} for all classification tasks.}
    \label{fig:cls}
\end{figure}

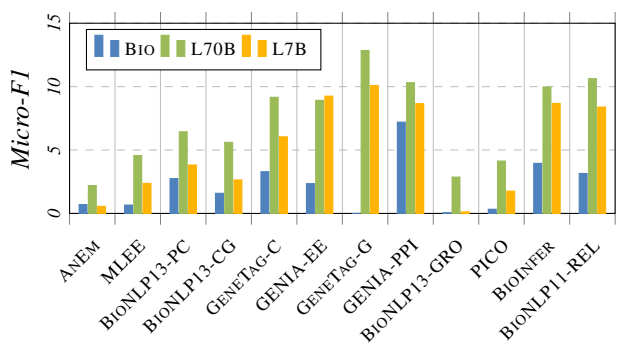
\begin{figure}[t]
\centering
    \begin{tikzpicture}
    \begin{axis}[
        ybar=0.5pt,
        bar width=3pt,
        x tick label style={rotate=45, font=\scriptsize\scshape, anchor=north east},
        x tick label as interval,
        y tick label style={rotate=90, font=\scriptsize\itshape},
        height=11.7em,
ylabel=\emph{Micro-F1},
        width  = 1.05\columnwidth,
        ymajorgrids=true,
        y grid style=dashed,
        major y tick style = transparent,
        xtick={0, 1, 2, 3, 4, 5, 6, 7, 8, 9, 10, 11, 12},
        ytick={0, 0.05, 0.10, 0.15},
        yticklabels={0, 5, 10, 15},
        xmajorgrids=true,
        xminorgrids=true,
        legend pos=north west,
        legend cell align={left},
        legend columns=3,
        xmin=0, xmax=12,
        ymin=0.0, ymax=0.15,
        xticklabels={AnEm, MLEE, BioNLP13-PC, BioNLP13-CG, GeneTag-C, GENIA-EE,  GeneTag-G, GENIA-PPI, BioNLP13-GRO, PICO, BioInfer, BioNLP11-REL}
    ]
    \addplot[legend entry=\textsc{\scriptsize Bio}, color=bblue, fill=bblue]  coordinates {(0.5, 0.0070) (1.5, 0.0066) (2.5, 0.0275) (3.5, 0.0159) (4.5, 0.0329) (6.5, 0.0000) (5.5, 0.0235) (7.5, 0.0720) (8.5, 0.0005) (9.5, 0.0033) (10.5, 0.0395) (11.5, 0.0315)};
    \addplot[legend entry=\textsc{\scriptsize L70B}, color=ggreen, fill=ggreen]  coordinates { (0.5, 0.0220) (1.5, 0.0456) (2.5, 0.0644) (3.5, 0.0561) (4.5, 0.0916) (6.5, 0.1285) (5.5, 0.0892) (7.5, 0.1031) (8.5, 0.0287) (9.5, 0.0413) (10.5, 0.0997) (11.5, 0.1063)};
    \addplot[legend entry=\textsc{\scriptsize L7B}, color=rred, fill=rred]  coordinates {(0.5, 0.0056) (1.5, 0.0237) (2.5, 0.0382) (3.5, 0.0264) (4.5, 0.0604) (6.5, 0.1009) (5.5, 0.0925) (7.5, 0.0866) (8.5, 0.0012) (9.5, 0.0176) (10.5, 0.0868) (11.5, 0.0839)};
    \end{axis}
    \end{tikzpicture}
    \vspace{-1\baselineskip}
    
    \caption{Best-performing \emph{Standard Prompting} method for \textcolor{bblue}{\bfseries Bio}Mistral 7B, \textcolor{ggreen}{\bfseries L}lama-\textcolor{ggreen}{\bfseries 70B} and \textcolor{rred}{\bfseries L}lama-\textcolor{rred}{\bfseries 7B} for all NER tasks.}
    \label{fig:generalisation-mcq-2}
\end{figure}

\textbf{Reasoning and knowledge-enhancing techniques do not improve performance.} Figures~\ref{fig:cls}~and~\ref{fig:generalisation-mcq-2} compare the best-performing techniques for classification and NER. Surprisingly, Table~\ref{table:combined_results_with_span} in the Appendix shows that Standard Prompting consistently achieves the highest average F1 scores across models: BioMistral-7B (36.48\%), Llama-2-70B-Chat-AWQ (40.34\%), and Llama-2-7b-chat-hf (34.92\%). This suggests that for structured prediction tasks, complex reasoning techniques like CoT or RAG do not outperform Standard Prompting.

For NER, Standard Prompting remains effective, but performance varies across models and datasets. Scores are significantly lower than typical F1 scores in biomedical NER benchmarks such as NCBI disease corpus~\citep{dougan2014ncbi, krallinger2015chemdner} and CHEMDNER, where specialized models achieve up to 0.90 Span F1 scores~\citep{kocaman2021biomedical, zhou2023universalner}. However, similar to our findings, zero-shot NER scores tend to be low, even in general domains~\cite{shen2021locate} and when providing label descriptions~\cite{DBLP:journals/corr/abs-2406-02245}.

The likely reason for poor performance is that these approaches excel in knowledge- and reasoning-intensive tasks like Question Answering~\cite{DBLP:journals/corr/abs-2311-16452} or Mathematical Reasoning~\cite{wang2024chain, wang2022self, li2024rt}, but structured prediction tasks require understanding task semantics over generic reasoning. These tasks rely less on broad knowledge from biomedical papers or Wikipedia and more on domain-specific application within the given input. Effective models must handle specialized vocabulary, jargon, acronyms, and synonyms varying across subfields~\cite{kim2007beyond, zheng2018assessing, jiang2024medreadme}. They must also resolve ambiguity, polysemy, and syntactic nuances in biomedical concepts, which the LLMs to not have been able to capture.

\textbf{Scale drives improvements.} Consistent with prior findings, the 70B model shows notable gains over the 7B model (5.4\% for classification, 2.2\% for NER Span F1). The largest performance gap appears when using SC with COT and RAG (Wikipedia), where the 70B model surpasses the 7B model by 15.45\%. This suggests the larger model excels at leveraging external knowledge when paired with SC and chain-of-thought prompting.
The 70B model's greater capacity is particularly beneficial for handling complex reasoning and knowledge integration~\cite{wei2022chain}. This is further supported by its 10.91\% improvement when SC is added to Wikipedia-based RAG, helping mitigate performance drops from irrelevant external information. Unlike classification tasks, where Standard Prompting performed best, NER performance remains stable with advanced prompting techniques, especially in larger models like Llama-2-70B, likely due to the inherent lack of epistemic certainty in NER outputs.


\subsection{Detailed Comparison of Prompting Techniques}
\begin{figure}  
    \begin{tikzpicture}
    \begin{axis}[
        xbar=0.5pt,
        bar width=3pt,
        y tick label style={font=\scriptsize\scshape},
        x tick label style={font=\scriptsize\itshape},
        xtick={0, 0.2, 0.4, 0.6},
        xticklabels={0, 0.2, 0.4, 0.6},
        height=0.40\columnwidth,
        width=0.95\columnwidth,
        xlabel=\emph{F1 Score},
        xmajorgrids=true,
        x grid style=dashed,
        major x tick style = transparent,
        ytick={0, 1, 2},
        yticklabels={L70B, L7B, Bio},
        legend pos=north east,
        legend cell align={left},
        legend style={font=\scriptsize},
        xmin=0, xmax=0.69,
        ymin=-0.5, ymax=2.5,
        rotate=0,
        legend columns=1
    ]
    \addplot[mark=none, rred, dashed, sharp plot] coordinates {(0.215, -1) (0.215, 3)};
    \addplot[mark=none, bblue, dashed, sharp plot] coordinates {(0.415, -1) (0.415, 3)};
    \addplot[legend entry=Single-Class, color=bblue, fill=bblue] coordinates {
        (0.42, 0) 
        (0.36, 1) 
        (0.35, 2) 
    };
    \addplot[legend entry=Multi-Class, color=rred, fill=rred] coordinates {
        (0.17, 0) 
        (0.16, 1) 
        (0.13, 2) 
    };
    \end{axis}
    \end{tikzpicture}
    \caption{Performance comparison for {\bfseries Bio}Mistral 7B, {\bfseries L}lama{\bfseries 7B} and {\bfseries L}lama{\bfseries 70B} on single- and multi-label datasets, with random guess baselines of 0.415 and 0.215, respectively.}
    \label{fig:cls_single_vs_multi}
\end{figure}

\textbf{CoT and SC underperform without sufficient parametric knowledge.} For BioMistral-7B, SC-CoT prompts reduce classification performance by about 16\%. One reason may be that domain-specific pre-training, while enhancing performance on specialized biomedical tasks, can limit the model’s general adaptability~\cite{brokman2024important}. Similar to RAG, SC does not consistently improve NER. While SC aims to generate multiple reasoning paths and select the most consistent one, it may introduce errors when the model's epistemic certainty in its outputs is low, leading to performance drops. For NER, combining CoT and SC with RAG (Wikipedia) produces the largest performance gap between 70B and 7B models, suggesting that larger models use external knowledge and complex reasoning more effectively when parametric knowledge is limited.

\textbf{RAG does not help information extraction.} Although RAG can improve QA tasks by retrieving relevant facts~\cite{xiong2024benchmarking}, it seems less useful for classification and information extraction, where irrelevant information misleads the model, creating additional complexity. This results in a drop in classification accuracy, dropping 16.91\% with PubMed Corpora and 16.47\% with Wikipedia compared to the best classification method.

\textbf{SC helps filter noise for RAG but does not aid CoT.} While SC aims to improve CoT by generating multiple reasoning paths, its efficacy depends on the model's epistemic certainty~\cite{yadkori2024believe, liu2024uncertainty}. Insufficient parametric knowledge lowers confidence, causing performance declines. BioMistral-7B loses about 16\% in classification with SC-CoT prompting. SC also fails to consistently boost NER. However, in the 70B model, combining CoT and SC with RAG (Wikipedia) yields notable gains, indicating that larger models better exploit external knowledge and present higher epistemic certainty owing to their larger parametric capacity.

\subsection{Detailed Per-Dataset Analysis}
Figure \ref{fig:cls_single_vs_multi} provides the performance comparison of the three models on single and multi-label tasks. Figures \ref{fig:cls-by-dataset} and \ref{fig:ner-by-dataset} provide a detailed analysis and breakdown of performance of each technique (NER and classification) on each dataset, along with random baselines for each. We discuss their implications below.

\label{sec:per-dataset-analysis}
\textbf{Models perform significantly better on public datasets.} On public datasets, models average around 30\% accuracy, compared to 12\% on private datasets, suggesting potential data leakage from publicly available sources used in pre-training or instruction-tuning. Thus, performance on “unseen” tasks may stem from prior exposure rather than true generalization.

\textbf{Multilingual performance is not scale dependent.} As shown in Figure~\ref{fig:cls}, smaller models can match or outperform larger ones on Chinese and Japanese datasets but lag behind in English tasks. This disparity likely results from heavy reliance on English corpora and limited non-English domain exposure, increasing overfitting risks. Factors like language family, data sampling ratios, and sparse representation can also inhibit multilingual models~\cite{he2024scaling,bagheri2024drives}.

\textbf{LLMs struggle on high-complexity tasks.} As Figure~\ref{fig:cls_single_vs_multi} shows, on average, the models fail to surpass random baselines for single and multi-class classification, though Figure~\ref{fig:cls-by-dataset} shows these baselines remain unbeaten in only two of 14 datasets. 

\section{Conclusion}

We provide a comprehensive benchmark and analysis of LLMs in Medical Classification and Named Entity Recognition tasks, revealing several key insights that have significant implications for the field. 
We carry out a critical investigation of broad claims regarding LLM capabilities by replicating them in various contexts, domains and datasets. 
We find that models suffer from fundamental drawbacks in generalizability, which hinder their performance in structured information extraction tasks on domain specific problems. 
This leads to standard prompting outperforming more advanced methods for all models.
Our findings underscore that parametric knowledge capacity remains the most important factor in zero-shot settings, with advanced techniques used to augment external knowledge or model reasoning failing to improve performance.

\clearpage
\section*{Acknowledgement}
Viktor Schlegel is part of the IN-CYPHER programme and is supported by the National Research Foundation, Prime Minister’s Office, Singapore under its Campus for Research Excellence and Technological Enterprise (CREATE) programme. We are grateful for the support provided by Research IT to him in form of access to the Computational Shared Facility at The University of Manchester and the computational facilities at the Imperial College Research Computing Service\footnote{DOI: \url{https://doi.org/10.14469/hpc/2232}}.

\section*{Limitations}
While our study provides important insights into LLMs' capabilities for biomedical classification and information extraction, several limitations should be considered when interpreting our results. Our findings are primarily empirical and, although they suggest consistent patterns across models and tasks, further theoretical work is needed to fully understand why advanced prompting techniques fail to improve performance on structured prediction tasks. We deliberately exclude closed-source LLMs from our analysis due to privacy concerns in medical applications and the observed dataset leakage issues, where public dataset contamination is even harder to control for proprietary models. Additionally, our focus on constrained decoding for reliable output parsing necessarily limits us to open-source models where we have access to the generation process.

We also specifically choose not to evaluate in-context learning (ICL) approaches, as our study focuses on ``true'' zero-shot capabilities where no task-specific examples are available. While techniques like k-NN ICL have shown promise in other domains, they require substantial annotated data to retrieve examples from---which is often unavailable in practical medical settings. Fixed ICL examples could be used, but performance would then largely depend on example selection, essentially reducing the evaluation to the quality of prompt engineering. To balance \emph{(i)} scientific validity and focus on real-world scenarios, where domain experts may not be prompt engineering specialists, with \emph{(ii)} the need to provide useful information to the models, we instead opt for the zero-shot setting---addressing \emph{(i)}---while ensuring semantic clarity through meaningful label names (e.g., using ``Control'' and ``Perturbation'' rather than ``0'' and ``1'' in the \textsc{GEO} dataset)---addressing~\emph{(ii)}.








\bibliography{references,references_2, references_3}

\clearpage

\appendix

\section{Related Work}
\label{sec:appendix-rel-work}






\textbf{Existing LLMs Benchmarks:} With the rising popularity of LLMs, many works evaluated their performance in the biomedical and clinical domains.
These works typically focus on evaluating domain-knowledge by means of Question Answering~\cite{singhal2023large, harris2023large,Subramanian2024M-QALM:Answering}, or focus directly on possible application scenarios, such as summarisation \cite{DBLP:conf/bionlp/LiWSBNKZB0N23, DBLP:journals/corr/abs-2306-02022} or clinical coding \cite{DBLP:journals/eswa/KaurGO23}. Many works combine these two directions in an effort to provide more comprehensive benchmarks \cite{srivastava2024functional,xiong2024benchmarking,feng2024evaluation, DBLP:journals/corr/abs-2004-03329, DBLP:journals/corr/abs-2401-14493}. However, many of these works overlook the wealth of existing literature and plethora of available resources for traditional structured prediction tasks in the biomedical domain, such as document classification, entity recognition and linking and event and relation extraction (e.g., \citeauthor{pyysalo2012event} (\citeyear{, pyysalo2007bioinfer,pyysalo2012event}) to name a few). \citet{fries2022bigbio} have provided a comprehensive and unified collection of these resources, however their work prioritises reportage of the resource collection over benchmarking results. Their preliminary evaluations suggest that their evaluated pre-LLM era models barely surpass the random guess baseline in the zero-shot setting. We build upon their work by providing a detailed analysis to what extent approaches to enhance reasoning and knowledge in LLMs help to challenge this status quo.

\section{Datasets}
Table \ref{tab:classification} and \ref{tab:ner} list the huggingface dataset cards and citations for each classification and ner dataset used in the paper respectively. \\For datasets considered private, we assume that models have not been trained on these datasets due to their restricted access, which requires Data Use Agreements (DUAs) and other permissions. Consequently, the likelihood of these datasets being included in common web crawls is low. \\ We have signed all the relevant Data Use Agreements (DUAs) and strictly adhere to their provisions. We do not redistribute the data and advise those wishing to reproduce experiments involving private datasets to consult the corresponding Hugging Face dataset cards for guidance on obtaining the necessary data.

\section{Compute Details}
\begin{enumerate}
    \item Hardware used (GPU/CPU): We used a mix of different shared computational facilities with nVidia A100-SXM4-80GB, RTX6000 with 24GB and L40S with 48GB. Debian OS was used for all the compute servers.
    \item Memory: The machines used had between 256 GB and 1TB of memory
    \item Software and libraries used: The environment can be reproduced from the 
    texttt{environment.yaml} file in the supplementary material
    \item Model details: The models used have been described in detail in the main paper submission under the Models subsection of the Methodology section. 
    \item Random seed of 42 was used for all random sampling purposes
\end{enumerate}

\begin{table*}[t]
\centering
\begin{tabularx}{\textwidth}{|X|X|X|}
\hline
\textbf{Dataset Name} & \textbf{HuggingFace Card} & \textbf{Citation} \\ \hline
\textsc{GAD} & bigbio/gad & \cite{Bravo2015} \\ \hline
\textsc{GEO} & bigbio/geokhoj\_v1 & \cite{geokhoj_v1} \\ \hline
\textsc{MedDialog} & bigbio/meddialog & \cite{DBLP:journals/corr/abs-2004-03329} \\ \hline
\textsc{CZIBase} & bigbio/czi\_drsm & \\ \hline
\textsc{CZIQoL} & bigbio/czi\_drsm & \\ \hline
\textsc{CZINatHist} & bigbio/czi\_drsm & \\ \hline
\textsc{LitCovid} & bigbio/bc7\_litcovid & \cite{chen2021overview} \\ \hline
\textsc{CAS} & bigbio/cas & \cite{grabar-etal-2018-cas} \\ \hline
\textsc{ESSAI} & bigbio/essai & \cite{grabar-etal-2018-cas} \\ \hline
\textsc{NTCIR13-Ja} & \makecell[l]{bigbio/ntcir\_13\\\_medweb} & \cite{iso2017ntcir13} \\ \hline
\textsc{NTCIR13-En} & \makecell[l]{bigbio/ntcir\_13\\\_medweb} & \cite{iso2017ntcir13} \\ \hline
\textsc{NTCIR13-Zh} & \makecell[l]{bigbio/ntcir\_13\\\_medweb} & \cite{iso2017ntcir13} \\ \hline
\textsc{PsyTAR} & bigbio/psytar & \cite{Zolnoori2019} \\ \hline
\textsc{SciCite} & bigbio/scicite & \cite{cohan:naacl19} \\ \hline
\end{tabularx}
\caption{Datasets used for classification tasks.}
\label{tab:classification}
\end{table*}

\begin{table*}[t]
\centering
\begin{tabularx}{\textwidth}{|X|X|X|}
\hline
\textbf{Dataset Name} & \textbf{HuggingFace Card} & \textbf{Citation} \\ \hline
\textsc{GeneTag-G} & bigbio/genetag & \cite{Tanabe2005} \\ \hline
\textsc{GeneTag-C} & bigbio/genetag & \cite{Tanabe2005} \\ \hline
\textsc{GENIA-PPI} & \makecell[l]{bigbio/genia\\\_relation\_corpus} & \cite{pyysalo-etal-2009-static, Hoehndorf_applyingontology, article} \\ \hline
\textsc{AnEm} & \makecell[l]{bigbio/an\_em} & \cite{ohta-etal-2012-open} \\ \hline
\textsc{BioInfer} & \makecell[l]{bigbio/bioinfer} & \cite{pyysalo2007bioinfer} \\ \hline
\textsc{Genia-EE} & \makecell[l]{bigbio/bionlp\\\_shared\_task\_2009} & \cite{kim-etal-2009-overview} \\ \hline
\textsc{BioNLP11-REL} & \makecell[l]{bigbio/bionlp\_st\\\_2011\_rel} & \cite{10.5555/2107691.2107703} \\ \hline
\textsc{BioNLP-13-CG} & \makecell[l]{bigbio/bionlp\_st\\\_2013\_cg} & \cite{pyysalo-etal-2013-overview} \\ \hline
\textsc{BioNLP-13-GRO} & \makecell[l]{bigbio/bionlp\_st\\\_2013\_gro} & \cite{kim-etal-2013-gro} \\ \hline
\textsc{BioNLP-13-PC} & \makecell[l]{bigbio/bionlp\_st\\\_2013\_pc} & \cite{ohta-etal-2013-overview} \\ \hline
\textsc{PICO} & \makecell[l]{bigbio/ebm\_pico} & \cite{nye-etal-2018-corpus} \\ \hline
\textsc{MLEE} & \makecell[l]{bigbio/mlee} & \cite{pyysalo2012event} \\ \hline
\end{tabularx}
\caption{Datasets used for NER tasks.}
\label{tab:ner}
\end{table*}


\section{Methodology}
\label{sec:techniques-full}

\paragraph{Datasets}

Since we evaluate different prompting techniques, we restrict the choice of tasks to those where the number of possible labels is small enough to fit in the evaluated LLMs' context window. 
We restrict the number of labels to ten and the mean length of the input documents to at most 2048 tokens. This leaves us with 14 different classification datasets from the BigBio collection\footnote{for the GAD dataset, we only select 1 fold out of the 10 available, as the folds feature the same task for different data, unlike other datasets. We also skipped the Chinese subset of meddialog as we had difficulties loading the dataset}. For the NER task, we sample 12 datasets from the pool of those that satisfy the criteria. The resulting dataset sample features four non-English datasets and six non-public classification datasets, which allows us to investigate whether LLMs perform better on minority languages or on data that is less likely to be found in public pre-training corpora.
We run the evaluation on the official test-set split where available, otherwise we consider the full dataset. For datasets with more than 500 instances, we sample 500 random but fixed instances to speed up the experiments. Overall, our selection spans English and non-english source data, publicly available and private datasets, and various domains such as scientific papers, medical notes and social media. 

\emph{Classification:}
The datasets used for classification tasks include both single-label and multi-label datasets, covering a wide range of biomedical and clinical domains. For single-label classification, the \textsc{GAD} dataset focuses on identifying associations between genes and diseases \cite{Bravo2015}, while the \textsc{GEO} dataset is concerned with classifying microarray, transcriptomics, and single-cell experiments from the Gene Expression Omnibus (GEO) database \cite{geokhoj_v1}. The \textsc{MedDialog} dataset aims to classify dialogue snippets as either being said by a doctor or a patient \cite{DBLP:journals/corr/abs-2004-03329}. Furthermore, the \textsc{CZIDrsm} dataset has several subsets, including one for classifying research articles based on aspects of disease research (\textsc{CZIBase}), and others for identifying whether a paper describes substantive research into Quality of Life (\textsc{CZIQoL})  or is a natural history study (\textsc{CZINatHist}).

In multi-label classification, the \textsc{LitCovid} dataset is used for the classification of COVID-19-related articles \cite{chen2021overview}. The \textsc{CAS} and \textsc{ESSAI} datasets are utilized for identify negation and uncertainty clinical cases from French-speaking countries \cite{grabar-etal-2018-cas}. The \textsc{NTCIR13} datasets include subsets for disease classification of tweets in Japanese (\textsc{*-Ja}), English (\textsc{*-En}), and Chinese (\textsc{*-Zh}) \cite{iso2017ntcir13}. Additionally, the \textsc{PsyTAR} dataset is used for sentence classification of various drug-related effects, such as Adverse Drug Reactions (ADR) and Withdrawal Symptoms (WDs) \cite{Zolnoori2019}, while the \textsc{SciCite} dataset is used for citation intent classification based on the context within computer science and biomedical domains \cite{cohan:naacl19}.

\emph{NER:}
The datasets for Named Entity Recognition (NER) tasks are similarly divided into entity recognition (single entity type) and classification (multiple entity types). In the single-type category, the \textsc{GeneTag} dataset is used for gene/protein NER, with two annotation versions: the original \textsc{GeneTag-G}  and the corrected \textsc{GeneTag-C}  \cite{Tanabe2005}. Additionally, the \textsc{GENIA-PPI} dataset focuses on protein-protein interactions or gene regulatory relations within the GENIA corpus, capturing primarily static relations \cite{pyysalo-etal-2009-static, Hoehndorf_applyingontology, article}.

The multiple-type NER datasets encompass various complex biomedical tasks. The \textsc{AnEm} dataset targets anatomical entity recognition \cite{ohta-etal-2012-open}, while the \textsc{BioInfer} dataset focuses on recognizing proteins, genes, and RNA entities \cite{pyysalo2007bioinfer}. The \textsc{Genia-EE} dataset is used for the GENIA Event corpus \cite{kim-etal-2009-overview}, and the \textsc{BioNLP11-REL} dataset is employed for extracting part-of relations between genes/proteins and associated entities~\cite{10.5555/2107691.2107703}. Furthermore, the \textsc{BioNLP-13-CG} dataset is used for Cancer Genetics (CG) information extraction, focusing on recognizing events represented as structured n-ary associations of given physical entities \cite{pyysalo-etal-2013-overview}. The \textsc{BioNLP-13-GRO} dataset aims to populate the Gene Regulation Ontology with events and relations \cite{kim-etal-2013-gro}, and the \textsc{BioNLP-13-PC} dataset is used for the automatic extraction of biomolecular reactions from text \cite{ohta-etal-2013-overview}. Lastly, the \textsc{PICO} dataset deals with recognizing (P)articipants, (I)nterventions, and (O)utcomes \cite{nye-etal-2018-corpus}, and the \textsc{MLEE} dataset is used for event extraction related to angiogenesis \cite{pyysalo2012event}.

\paragraph{Models} For our experiments, we employed two instruction-tuned variants of the Llama-2 model---7B and 70B---both \cite{touvron2023llama}, alongside the BioMistral-7B model \cite{labrak2024biomistral} which was further pre-trained on the biomedical domain. 
Since we make use of constrained generation to generate model outputs and guide the models decoding process, we retrict the evaluation to open source models since this process is not possible for proprietary models such as GPT-4.

\paragraph{Techniques} Table~\ref{tab:techniques_summary} summarizes the techniques used in this study and highlights relevant nuances and comments. These techniques include \textsc{Vanilla} (standard prompting), \textsc{CoT} (chain-of-thought reasoning) \cite{wei2022chain}, and \textsc{SC} (self-consistency) \cite{wang2022self}, as well as \textsc{RAG} (retrieval-augmented generation) \cite{lewis2020retrieval}. For \textsc{RAG}, we used FAISS \cite{douze2024faiss, johnson2019billion} with PubMed abstracts \cite{DBLP:journals/jis/SanyalBD21} and Wikipedia articles as corpora, embedding documents with \texttt{all-MiniLM-L6-v2} \cite{DBLP:conf/emnlp/ReimersG19}. We also implemented constrained decoding for structured output generation \cite{willard2023efficient}, crucial for ensuring reliable outputs in NER and classification tasks. 
A novel two-stage approach for NER was adopted, inspired by \cite{shen2021locate}, where general entities were assigned in Stage 1 and refined in Stage~2. 

Standard prompting was used as a baseline for both the Classification as well as the NER tasks. 
\textit{Chain-of-thought reasoning} \cite{wei2022chain} has been shown to improve performance, particularly in QA and logical reasoning tasks. Thus, we also ran experiments with \textit{chain-of-thought} reasoning to measure its impact on model performance. For the NER task, we adapted a more guided, \textit{two-stage approach} \cite{shen2021locate} to implement a novel chain-of-thought reasoning approach. 
Here, The first stage involves inducing a generic entity name from a datasets' known entity labels---e.g., ``Bodypart'' for the NER labels describing different bodyparts---and then labelling the input document with that generic entity type.
In the second stage all entities labelled in this way are further disambiguated with their respective fine-grained dataset NER labels. 
\textit{Retrieval Augmented Generation} \cite{lewis2020retrieval} has been established as an effective technique to improve model performance by introducing relevant non-parameteric knowledge to models and thus grounding the generated outputs to factual information. \citet{xiong2024benchmarking} conducted a systematic study of RAG on medical QA, and we incorporate their findings 
into our study. 
We used PubMed abstracts \cite{DBLP:journals/jis/SanyalBD21} and Wikipedia articles as knowledge corpora,  because \citeauthor{xiong2024benchmarking}'s (\citeyear{xiong2024benchmarking}) experiments found that using PubMed improved performance over non RAG techniques, while using Wikipedia reduced performance in medical QA tasks. Our goal was to evaluate whether the same holds true for structured prediction tasks as well. 
For the RAG module, we made use of FAISS~\cite{douze2024faiss, johnson2019billion}, which allows retrieval of most similar documents based on semantic similarity, where we used the 
\texttt{all-MiniLM-L6-v2} sentence transformers \cite{DBLP:conf/emnlp/ReimersG19} model for embedding input documents and corpora. For each experiment, the number of retrieved documents was computed based on the maximum possible documents which could be used without exceeding the token limit of the model. \\ 
\textit{Self-consistency}, proposed by \citet{wang2022self}, improves chain-of-thought reasoning of LLMs by sampling reasoning paths for a given problem, followed by a majority vote  for the final answer. We also conduct a set of experiments employing \textit{self-consistency} to investigate whether such improvements can be observed on structured prediction tasks in the medical domain as well. For classification tasks, self consistency was employed to generate multiple reasoning chains for the given problem, followed by answer extraction from each reasoning chain and majority voting to select the final answer. For NER tasks, since we follow the two-stage approach, self-consistency was employed in both stages. Multiple general entity labels were generated in the first stage, and entities were extracted for each such label. In the second stage, self consistency was again used for the entity selection phase as well as the entity label determination step. Majority voting was utilised in final label or class selection in each case \cite{xie2023empirical}.\\
\textit{Constrained decoding} in LLMs~\cite{willard2023efficient} was used to ensure structured information extraction and text generation. This allowed us to evaluate the LLMs for the task at hand without the added variability due to the aleatoric uncertainties brought about by the probabilistic language generation fundamental to the architectures of the models. More specifically, for classification tasks, we ensured the presense of at least one label in the generated outputs. For NER we restricted the generation of spans occurring in text in the first step, and in the second step, for each of the spans we restricted the generation to any of the possible labels. 
This is also one of the reasons why we opted against evaluating API-based closed-source LLMs\footnote{The other reason being their intransparancy with regard to training data, which violates our ``true'' zero-shot setting.}, as in our initial experiments the hallucinations in generated outputs created problems with reliably parsing the structured outputs.

We refer to chain of thought as \textsc{CoT}, Self-consistency as \textsc{SC}, RAG as \mbox{\textsc{RAG-\{P\textbar W\}}} for PubMed and Wikipedia corpora, respectively, and to standard prompting as \textsc{Vanilla}.

\section{Analysis and Performance Breakdown}
\label{sec:appendix-analysis}

Figures \ref{fig:cls-by-dataset} and \ref{fig:ner-by-dataset} provide a detailed analysis and breakdown of performance of each technique (NER and classification) on each dataset, along with random baselines for each. Figure \ref{fig:cls_single_vs_multi} provides the performance comparison of the three models on single and multi-label tasks. A complete discussion for these figures and their implications can be found in section \ref{sec:per-dataset-analysis}.

As discussed in section \ref{sec:per-dataset-analysis}, LLMs struggle on high-complexity tasks. Even the best performing model, Llama2 70B performs well on only relatively low-complexity tasks (\textsc{CZIBase}, \textsc{NTCIR13-En}) and moderate tasks (\textsc{Geo}), but struggles with higher-complexity datasets (\textsc{BioNLP13-CG}, \textsc{GENIA-EE}). 
In tasks requiring nuanced interpretation (\textsc{PICO}, \textsc{BioNLP13-GRO}), performance remains low. Although RAG (Retrieval-Augmented Generation) sometimes boosts results, it does not universally enhance biomedical information extraction or classification. These findings indicate that even the most advanced general-purpose and domain-specific LLMs are not good zero-shot reasoners for structured prediction tasks such as biomedical information extraction, especially for complex task settings.

\section{Results Analysis}

\begin{figure*}[!tbh]
\centering
    \begin{tikzpicture}
    \begin{axis}[
        ybar=0.5pt,
        bar width=2.5pt,
        x tick label style={rotate=45, font=\scriptsize\scshape, anchor=north east},
        x tick label as interval,
        y tick label style={rotate=90, font=\scriptsize\itshape},
        ytick={0, 0.2, 0.4, 0.6, 0.8},
        yticklabels={0, 20, 40, 60, 80},
        height=14.7em,
ylabel=\emph{Micro-F1},
        width  = 1.05\textwidth,
        ymajorgrids=true,
        y grid style=dashed,
        major y tick style = transparent,
        xtick={0, 1, 2, 3, 4, 5, 6, 7, 8, 9, 10, 11, 12, 13, 14},
        xmajorgrids=true,
        xminorgrids=true,
        legend pos=north west,
        legend cell align={left},
        legend columns=4,
        xmin=0, xmax=14,
        ymin=0.0, ymax=0.86,
        xticklabels={CAS, SciCite, NTCIR13-Zh, NTCIR13-En, NTCIR13-Ja, CZIBase, ESSAI, GEO, GDA, LitCovid, MedDialog, CZiNatHist, PsyTAR, CZIQoL}
    ]
    \addplot[red, line legend,sharp plot,nodes near coords={},
    update limits=false,shorten >=-3mm,shorten <=-3mm] 
    coordinates {(0, 0.1818) (0.75, 0.1818)};
    \addplot[red, line legend,sharp plot,nodes near coords={},
    update limits=false,shorten >=-3mm,shorten <=-3mm] 
    coordinates {(1.25, 0.4040) (1.75, 0.4040)};
    \addplot[red, line legend,sharp plot,nodes near coords={},
    update limits=false,shorten >=-3mm,shorten <=-3mm] 
    coordinates {(2.25, 0.1520) (2.75, 0.1520)};
    \addplot[red, line legend,sharp plot,nodes near coords={},
    update limits=false,shorten >=-3mm,shorten <=-3mm] 
    coordinates {(3.25, 0.1520) (3.75, 0.1520)};
    \addplot[red, line legend,sharp plot,nodes near coords={},
    update limits=false,shorten >=-3mm,shorten <=-3mm] 
    coordinates {(4.25, 0.1520) (4.75, 0.1520)};
    \addplot[red, line legend,sharp plot,nodes near coords={},
    update limits=false,shorten >=-3mm,shorten <=-3mm] 
    coordinates {(5.25, 0.1840) (5.75, 0.1840)};
    \addplot[red, line legend,sharp plot,nodes near coords={},
    update limits=false,shorten >=-3mm,shorten <=-3mm] 
    coordinates {(6.25, 0.0896) (6.75, 0.0896)};
    \addplot[red, line legend,sharp plot,nodes near coords={},
    update limits=false,shorten >=-3mm,shorten <=-3mm] 
    coordinates {(7.25, 0.6820) (7.75, 0.6820)};
    \addplot[red, line legend,sharp plot,nodes near coords={},
    update limits=false,shorten >=-3mm,shorten <=-3mm] 
    coordinates {(8.25, 0.4880) (8.75, 0.4880)};
    \addplot[red, line legend,sharp plot,nodes near coords={},
    update limits=false,shorten >=-3mm,shorten <=-3mm] 
    coordinates {(9.25, 0.3251) (9.75, 0.3251)};
    \addplot[red, line legend,sharp plot,nodes near coords={},
    update limits=false,shorten >=-3mm,shorten <=-3mm] 
    coordinates {(10.25, 0.5246) (10.75, 0.5246)};
    \addplot[red, line legend,sharp plot,nodes near coords={},
    update limits=false,shorten >=-3mm,shorten <=-3mm] 
    coordinates {(11.25, 0.3020) (11.75, 0.3020)};
    \addplot[red, line legend,sharp plot,nodes near coords={},
    update limits=false,shorten >=-3mm,shorten <=-3mm] 
    coordinates {(12.25, 0.2621) (12.75, 0.2621)};
    \addplot[legend entry=\textsc{\scriptsize Guess}, color=red, fill=red, sharp plot]  coordinates {(13, 0.3080) (14, 0.3080)};

    \addplot[legend entry=\textsc{\scriptsize CoT}, color=bblue, fill=bblue]  coordinates {(0.5, 0.2722) (1.5, 0.3827) (2.5, 0.3333) (3.5, 0.3942) (4.5, 0.3488) (5.5, 0.3260) (6.5, 0.2102) (7.5, 0.6740) (8.5, 0.4440) (9.5, 0.2064) (10.5, 0.5984) (11.5, 0.3120) (12.5, 0.2641) (13.5, 0.2520)};
    \addplot[legend entry=\textsc{\scriptsize CoT-RAG-P}, color=ggreen, fill=ggreen]  coordinates {(0.5, 0.0571) (1.5, 0.0301) (2.5, 0.0000) (3.5, 0.0201) (4.5, 0.0043) (5.5, 0.3460) (6.5, 0.0500) (7.5, 0.4660) (8.5, 0.5100) (9.5, 0.0055) (10.5, 0.6230) (11.5, 0.1000) (12.5, 0.0030) (13.5, 0.0980)};
    \addplot[legend entry=\textsc{\scriptsize CoT-RAG-W}, color=rred, fill=rred]  coordinates {(0.5, 0.0000) (1.5, 0.0395) (2.5, 0.0170) (3.5, 0.0202) (4.5, 0.0043) (5.5, 0.3060) (6.5, 0.0175) (7.5, 0.5020) (8.5, 0.4860) (9.5, 0.0111) (10.5, 0.6393) (11.5, 0.0880) (12.5, 0.0060) (13.5, 0.0540)};
    \addplot[legend entry=\textsc{\scriptsize SC-CoT}, color=pink, fill=pink]  coordinates {(0.5, 0.2492) (1.5, 0.3062) (2.5, 0.0000) (3.5, 0.4466) (4.5, 0.0000) (5.5, 0.2820) (6.5, 0.2236) (7.5, 0.5860) (8.5, 0.5000) (9.5, 0.1543) (10.5, 0.5902) (11.5, 0.2660) (12.5, 0.0358) (13.5, 0.2820)};
    \addplot[legend entry=\textsc{\scriptsize SC-CoT-RAG-P}, color=teal, fill=teal]  coordinates {(0.5, 0.2021) (1.5, 0.3004) (2.5, 0.0000) (3.5, 0.4655) (4.5, 0.0000) (5.5, 0.2600) (6.5, 0.1764) (7.5, 0.5780) (8.5, 0.4980) (9.5, 0.1500) (10.5, 0.5902) (11.5, 0.2640) (12.5, 0.0514) (13.5, 0.2740)};
    \addplot[legend entry=\textsc{\scriptsize SC-CoT-RAG-P}, color=purple, fill=purple]  coordinates {(0.5, 0.2337) (1.5, 0.3113) (2.5, 0.0000) (3.5, 0.4475) (4.5, 0.0000) (5.5, 0.2700) (6.5, 0.1851) (7.5, 0.5160) (8.5, 0.4840) (9.5, 0.1754) (10.5, 0.5738) (11.5, 0.2540) (12.5, 0.0547) (13.5, 0.2140)};
    \addplot[legend entry=\textsc{\scriptsize Vanilla}, color=igreen, fill=igreen]  coordinates {(0.5, 0.3129) (1.5, 0.3270) (2.5, 0.3485) (3.5, 0.5849) (4.5, 0.4133) (5.5, 0.4140) (6.5, 0.2134) (7.5, 0.8560) (8.5, 0.4580) (9.5, 0.0934) (10.5, 0.6721) (11.5, 0.3840) (12.5, 0.2423) (13.5, 0.3280)};

    \end{axis}
    \end{tikzpicture}
    
    \caption{Breakdown of the Micro-F1 performance of each technique and the random guess baseline for all classification datasets, compared against the random guess baseline.}
    \label{fig:cls-by-dataset}
\end{figure*}

\begin{figure*}[!tbh]
\centering
    \begin{tikzpicture}
    \begin{axis}[
        ybar=0.5pt,
        bar width=3pt,
        x tick label style={rotate=45, font=\scriptsize\scshape, anchor=north east},
        x tick label as interval,
        y tick label style={rotate=90, font=\scriptsize\itshape},
        height=14.7em,
ylabel=\emph{Micro-F1},
        width  = 1.05\textwidth,
        ymajorgrids=true,
        y grid style=dashed,
        major y tick style = transparent,
        xtick={0, 1, 2, 3, 4, 5, 6, 7, 8, 9, 10, 11, 12},
        ytick={0, 0.05, 0.10, 0.15},
        yticklabels={0, 5, 10, 15},
        xmajorgrids=true,
        xminorgrids=true,
        legend pos=north west,
        legend cell align={left},
        legend columns=4,
        xmin=0, xmax=12,
        ymin=0.0, ymax=0.15,
        xticklabels={AnEm, MLEE, BioNLP13-PC, BioNLP13-CG, GeneTag-C, GENIA-EE,GeneTag-G, GENIA-PPI, BioNLP13-GRO, PICO, BioInfer, BioNLP11-REL}
    ]
    \addplot[legend entry=\textsc{\scriptsize Guess}, color=red, fill=red, sharp plot]  coordinates {(0, 0.00000) (1, 0.00000)};
    \addplot[color=red, fill=red, sharp plot]  coordinates {(1, 0.00000) (2, 0.00000)};
\addplot[color=red, fill=red, sharp plot]  coordinates {(2, 0.00139) (3, 0.00139)};
\addplot[color=red, fill=red, sharp plot]  coordinates {(3, 0.00013) (4, 0.00013)};
\addplot[color=red, fill=red, sharp plot]  coordinates {(4, 0.06567) (5, 0.06567)};
\addplot[color=red, fill=red, sharp plot]  coordinates {(5, 0.04119) (6, 0.04119)};
\addplot[color=red, fill=red, sharp plot]  coordinates {(6, 0.00215) (7, 0.00215)};
\addplot[color=red, fill=red, sharp plot]  coordinates {(7, 0.00214) (8, 0.00214)};
\addplot[color=red, fill=red, sharp plot]  coordinates {(8, 0.00002) (9, 0.00002)};
\addplot[color=red, fill=red, sharp plot]  coordinates {(9, 0.00010) (10, 0.00010)};
\addplot[color=red, fill=red, sharp plot]  coordinates {(10, 0.00293) (11, 0.00293)};
\addplot[color=red, fill=red, sharp plot]  coordinates {(11, 0.00214) (12, 0.00214)};

    \addplot[legend entry=\textsc{\scriptsize CoT}, color=bblue, fill=bblue]  coordinates {(0.5, 0.0220) (1.5, 0.0456) (2.5, 0.0644) (3.5, 0.0561) (4.5, 0.0916) (6.5, 0.1285) (5.5, 0.0892) (7.5, 0.1031) (8.5, 0.0287) (9.5, 0.0413) (10.5, 0.0997) (11.5, 0.1063)};
    \addplot[legend entry=\textsc{\scriptsize CoT-RAG-P}, color=ggreen, fill=ggreen]  coordinates {(0.5, 0.0481) (1.5, 0.0345) (2.5, 0.0658) (3.5, 0.0345) (4.5, 0.0114) (6.5, 0.0660) (5.5, 0.0692) (7.5, 0.1022) (8.5, 0.0105) (9.5, 0.0234) (10.5, 0.1010) (11.5, 0.0852)};
    \addplot[legend entry=\textsc{\scriptsize CoT-RAG-W}, color=rred, fill=rred]  coordinates {(0.5, 0.0677) (1.5, 0.0538) (2.5, 0.0245) (3.5, 0.0365) (4.5, 0.0760) (6.5, 0.1358) (5.5, 0.0910) (7.5, 0.1033) (8.5, 0.0228) (9.5, 0.0357) (10.5, 0.0992) (11.5, 0.1000)};
    \addplot[legend entry=\textsc{\scriptsize SC-CoT}, color=pink, fill=pink]  coordinates {(0.5, 0.0459) (1.5, 0.0343) (2.5, 0.0649) (3.5, 0.0341) (4.5, 0.0084) (6.5, 0.0210) (5.5, 0.0835) (7.5, 0.1114) (8.5, 0.0170) (9.5, 0.0308) (10.5, 0.1197) (11.5, 0.0813)};
    \addplot[legend entry=\textsc{\scriptsize SC-CoT-RAG-P}, color=teal, fill=teal]  coordinates {(0.5, 0.0503) (1.5, 0.0542) (2.5, 0.0154) (3.5, 0.0687) (4.5, 0.0741) (6.5, 0.1234) (5.5, 0.0874) (7.5, 0.1130) (8.5, 0.0172) (9.5, 0.0329) (10.5, 0.1222) (11.5, 0.1077)};
    \addplot[legend entry=\textsc{\scriptsize SC-CoT-RAG-P}, color=purple, fill=purple]  coordinates {(0.5, 0.0339) (1.5, 0.0383) (2.5, 0.0655) (3.5, 0.0283) (4.5, 0.0105) (6.5, 0.0351) (5.5, 0.0723) (7.5, 0.1046) (8.5, 0.0164) (9.5, 0.0274) (10.5, 0.1202) (11.5, 0.0853)};
    \addplot[legend entry=\textsc{\scriptsize Vanilla}, color=igreen, fill=igreen]  coordinates {(0.5, 0.0128) (1.5, 0.0746) (2.5, 0.0625) (3.5, 0.0422) (4.5, 0.0875) (6.5, 0.1120) (5.5, 0.0399) (7.5, 0.0710) (8.5, 0.0106) (9.5, 0.0295) (10.5, 0.0771) (11.5, 0.0792)};
    \end{axis}
    \end{tikzpicture}
    \vspace{-2\baselineskip}
    
    \caption{Breakdown of each technique and the random guess baseline on all NER datasets as measured by the Micro-F1 scores. A prediction is counted as correct when both the span and its assigned label are found in the ground truth}
    \label{fig:ner-by-dataset}
\end{figure*}

\end{document}